\documentclass[conference]{IEEEtran}
\IEEEoverridecommandlockouts
\usepackage{cite}
\usepackage{amsmath,amssymb,amsfonts}
\usepackage{algorithmic}
\usepackage{graphicx}
\usepackage{textcomp}
\usepackage{xcolor}
\usepackage{xspace}
\usepackage{xurl}
\def\BibTeX{{\rm B\kern-.05em{\sc i\kern-.025em b}\kern-.08em
 T\kern-.1667em\lower.7ex\hbox{E}\kern-.125emX}}
\begin{document}

\title{On-device AI: Quantization-aware Training of Transformers in Time-Series}

\author{
 \IEEEauthorblockN{Tianheng Ling \\ Supervised by Gregor Schiele}
 \IEEEauthorblockA{University of Duisburg-Essen, Germany}
 \IEEEauthorblockA{tianheng.ling@uni-due.de}
}

\maketitle
%============================================
\begin{abstract}
Artificial Intelligence (AI) models for time-series in pervasive computing keep getting larger and more complicated. The Transformer model is by far the most compelling of these AI models. However, it is difficult to obtain the desired performance when deploying such a massive model on a sensor device with limited resources. My research focuses on optimizing the Transformer model for time-series forecasting tasks. The optimized model will be deployed as hardware accelerators on embedded Field Programmable Gate Arrays (FPGAs). I will investigate the impact of applying Quantization-aware Training to the Transformer model to reduce its size and runtime memory footprint while maximizing the advantages of FPGAs.

\end{abstract}
%============================================


%============================================
\section{Introduction and Related Work}
Time-series analysis is an active application domain in pervasive computing. Recently, the success of the Transformer model \cite{vaswani2017attention} in  Natural Language Processing (NLP), demonstrating its potent modeling ability with time-series data, has piqued the interest of the time-series community. Most researchers in this community have concentrated on optimizing the architecture of the Transformer network to improve the forecasting of longer time-series sequences \cite{wen2022transformers}. Although relevant technologies for reducing model complexity have been proposed, the immense computational and memory requirements of the Transformer model make its deployment on end devices difficult. 

Quantization is a widely employed model compression approach that can reduce the size and computing intensity of full-precision (32-bit floating-point) data (such as weights, biases, and feature maps of a model) by representing them with fewer bits \cite{nagel2021white}. We have found that while quantization of Transformer models is well-established in NLP and Computer Vision, research on it in time-series is understudied.

Becnel et al.\cite{becnel2022tiny} applied 8-bit Post-Training Quantization (PTQ) to a Transformer model for an air pollution prediction task to fit an ESP32 microcontroller (MCU). One inference of the quantized Transformer model takes 176 milliseconds, too long for inference-sensitive applications. Similar to how GPUs can be used to increase the computational capability of CPUs, Field Programmable Gate Arrays (FPGA) can be used to enhance MCUs with specialized hardware accelerators. Implementing Transformer-specific accelerators on an embedded FPGA allows more complex computation and faster model inference.
Moreover, 8-bit quantization is insufficient to reduce the size of a complex model for resource-constrained devices. To enhance model performance in lower-bit quantization, we adopt Quantization-aware Training (QAT), which allows the model to learn while accounting for quantization errors during training\cite{nagel2021white}.

My research aims to obtain a low-bit, fully quantized Transformer model for time-series forecasting tasks. The quantized model will be implemented as a hardware accelerator on FPGAs to speed up model inference, supporting more complex applications. The following section outlines our team's published work to further clarify my motivation for doing this study. Then I describe my own work, including a problem formulation, a description of the full-precision Transformer model, and the current finished work on quantization implementation. I finish by discussing future research plans.

%=======================================================
\section{Published Work}
For the past five years, our team has been developing the \textit{Elastic Node}, an AI hardware platform dedicated to pervasive computing. At PerCom in 2018, we demonstrated power consumption measurements of a Multilayer Perceptron when inferring on the second generation of our hardware \cite{burger2018demo}. Then, we implemented a fixed-point quantized Convolutional Neural Network and performed a case study on an ECG classification task on the third generation \cite{burger2020embedded}. As the implementation of a Long Short-Term Memory Network was more complex, until 2022, we deployed it on the fourth generation and achieved higher energy efficiency than related work \cite{qianenhancing}. 

In a prior study, we used PTQ for quantization. This allowed us to reduce the number of quantization bits to 6 without performance degrading too much. In addition, we have developed the \textit{creator} toolchain \cite{chao2022creator} to generate accelerators on FPGAs automatically from deep learning models. However, this toolchain does not currently support Transformer networks. To extend the above work, my Ph.D. research starts with the study of ultra-low quantized Transformer models via QAT for conversion to accelerators on \textit{Elastic Node}'s FPGA.
%=======================================================
\section{On-going Research}

\subsection{Problem Formulation}
Forecasting univariate time-series is the task of my research. Future data points are to be predicted based on historical data points. The feature-label pairs for this supervised task are built using a sliding time window with a fixed length, i.e., time-series with $N$ data points ($x_{t-N+1}$, ..., $x_{t-1}$, $x_{t}$) for $M$-step ahead forecasting ($x_{t+1}$, ..., $x_{t+M-1}$, $x_{t+M}$). As a case study, I adopt a data set \cite{zhou2021informer} of power transformer oil temperature (OT). The OTs for the following 24 hours will be predicted based on the 48 known OTs. The data set is divided into a training set, a validation set, and a test set at a ratio of 3:1:1. Before input, each data set was zero-mean normalized with the same scaler. Mean Square Error (MSE) is used to evaluate the model's forecasting performance.

\subsection{Full-precision Transformer Model}
I successfully replicated the Transformer network developed by Wu et al. \cite{wu2020deep}, which enables single-step ahead uni-variate time-series forecasting using 48 previous time-series data points. By adding a position encoding operation to the decoder and applying the generative decoder proposed by Zhou et al. \cite{zhou2021informer}, I adapted the Transformer network to be capable of 24-step univariate time-series forecasting. This change enables the model to predict multiple unknown data points during a single forward propagation, resolving the issue where the decoder cannot execute parallel computation during inference. Compared to other models using the same data set, the performance of my Transformer is acceptable.

%=======================================================
\subsection{Quantization on Transformer}
Following a quantization analysis of this full-precision Transformer model, I initially investigated the applicability of linear quantization to all fully connected layers.

Firstly, I quantized only the weights and biases of these layers using a symmetric signed quantization scheme, where the valid range of these floating-point parameters in the tensor unit was determined via the Min-Max method. I found that quantizing the parameters of the model output layer can obviously diminish the model's predictive ability. Moreover, quantizing the parameters of the encoder's (or decoder's) input layer can have a small negligible impact on model performance and yields a limited gain in model compression. By retaining the parameters of these three linear layers as floating-point numbers, the Transformer model could be quantized to 2 bits without degrading performance, as long as the parameters of the linear layers of the context-attention module in the final decoder layer were not quantized, i.e., the attention distribution of this context-attention module was not destroyed. In this case,  I got the highest compression rate of 7.44498$\times$.

I further quantized the input of the quantizable linear layers using an asymmetric signed quantization scheme, where the efficient range of these floating-point inputs in the tensor unit was determined through the Moving Average Min-Max method. Despite introducing more quantization error, the performance of the 2-bit quantized model degraded by less than 10\%. As the quantization parameters of the input were also saved as part of the model parameters, the highest compression rate slightly decreased to 7.44361$\times$.

Moreover, considering that fixed-point matrix multiplication is faster and more resource-efficient than floating-point matrix multiplication on FPGAs, I optimized the inference for these linear layers. I theoretically validated with the PyTorch framework that quantizing the biases with the product of the input's scale factor and the weights' scale factor did not degrade model performance. The highest model compression rate increased to 7.46545$\times$. Given that the bit-shift operation is faster than the floating-point multiplication on FPGAs, I further approximated the product of the input's scale factor and the weights' scale factor with the right bit-shift operation. I theoretically confirmed that performing the full fixed-point operation in these linear layers did not obviously degrade the model performance. 

% ==============================================
\section{Future Work}
Upon the existing findings, I will explore mixed-precision quantization for better model compression rate, for example, by applying 16-bit quantization to the linear layers proven not to be quantizable while keeping the quantizable linear layers at 2-bit (or even 1-bit)
Then, I will quantize the remaining computations, such as the Self-Attention module, LayerNormalization layer, and Softmax function. After completing the full quantization of the model, I will assess the performance of the quantized model on the \textit{Elastic Node}'s FPGA.

% ==============================================
\section*{Acknowledgement}
The authors acknowledge the financial support provided by the Federal Ministry of Economic Affairs and Climate Action of Germany in the RIWWER project (01MD22007C).

% ==============================================
\bibliographystyle{IEEEtran}
\bibliography{reference}
\end{document}